\documentclass[10pt,twocolumn,letterpaper]{article}

\usepackage{booktabs} 
\usepackage{cvpr}
\usepackage{times}
\usepackage{epsfig}
\usepackage{graphicx}
\usepackage{amsmath,mathtools}
\usepackage{amssymb}
\usepackage{subcaption}

\usepackage{algorithm}
\usepackage{algpseudocode}
\algrenewcommand\algorithmicindent{.9em}%

\usepackage[pagebackref=true,breaklinks=true,letterpaper=true,colorlinks,bookmarks=false]{hyperref}

 \cvprfinalcopy 

\def\cvprPaperID{****} 

\ifcvprfinal\fi

\begin{document}

\title{Spot the Difference by Object Detection}

\author{Junhui Wu$^{1, 2}$\thanks{This work is done when Junhui Wu was an intern at JD Group} \qquad Yun Ye$^2$ \qquad Yu Chen$^2$ \qquad Zhi Weng$^2$ \\
$^1$ Tsinghua University \qquad $^2$ JD Group\\
{\tt\small wujh10@outlook.com, \{yeyun, chenyu6, wengzhi\}@jd.com}
}

\maketitle

\begin{abstract}
In this paper, we propose a simple yet effective solution to a change detection task that detects the difference between two images, which we call ``spot the difference''. Our approach uses CNN-based object detection by stacking two aligned images as input and considering the differences between the two images as objects to detect. An early-merging architecture is used as the backbone network. Our method is accurate, fast and robust while using very cheap annotation. 
We verify the proposed method on the task of change detection between the digital design and its photographic image of a book. Compared to verification based methods, our object detection based method outperforms other methods by a large margin and gives extra information of location. We compress the network and achieve 24 times acceleration while keeping the accuracy. Besides, as we synthesize the training data for detection using weakly labeled images, our method does not need expensive bounding box annotation.  	

\end{abstract}

\section{Introduction}
\label{sec:intro}
\begin{figure}
	\includegraphics[width=3.3in]{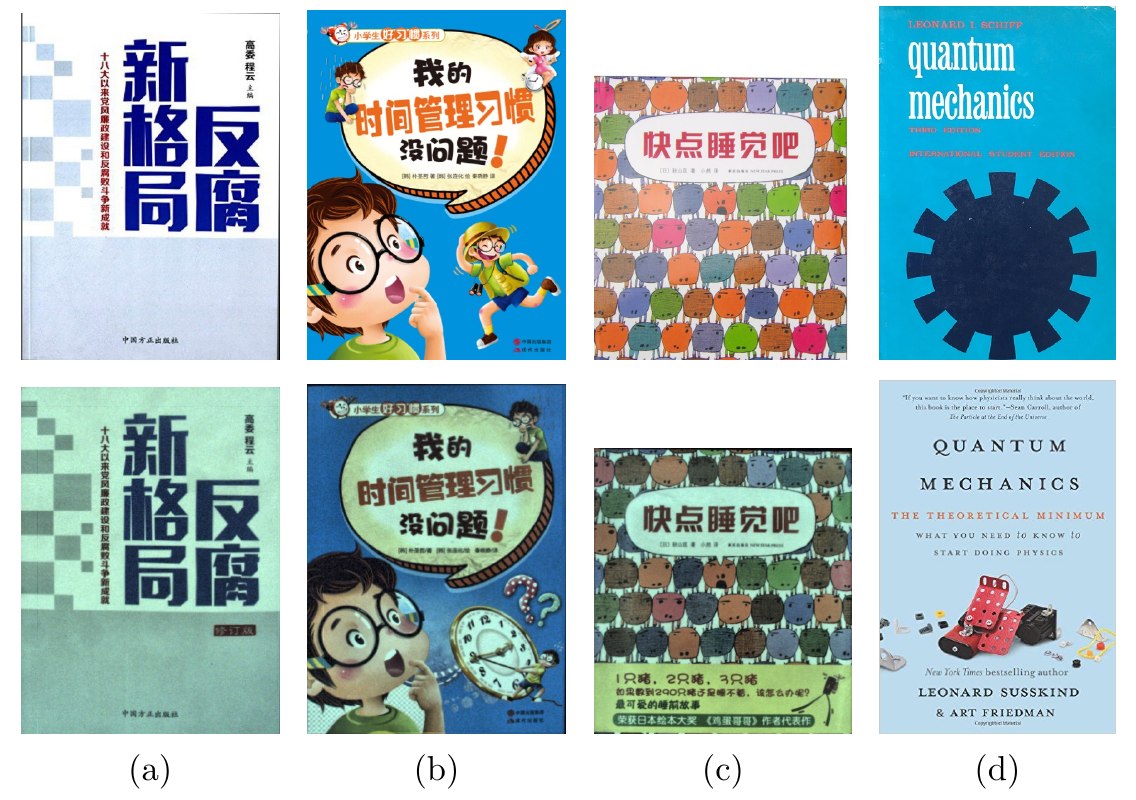}
	\caption{Some examples of changed book cover data. Digital designs are on the top and photographic images are on the bottom.  The differences are (a) text about version information, (b) local patterning, (c) book girdle, (d) the whole image respectively. The task is to spot these differences between digital design and photographic image.}\label{fig:realdata}
\end{figure}

This paper considers a problem of change detection between two book cover images, of which one is the digital design and the other is the photographic printed image.
Book cover digital design may somehow change at the time of publication, but the digital design displayed on the Internet or stored in database system is still the original one. 
Book sellers want to make sure that a printed book cover conforms to its digital design in the database system, because differences between the two images may cause commercial dispute.
In some cases, the book cover changes globally. In other cases, book cover changes locally but the change can not be ignored. For example, the first part and second part of one book series change locally, but the difference is very important. 
Some changed book cover photographic images and the original digital designs are shown in Figure ~\ref{fig:realdata}.

\begin{figure}
	\includegraphics[width=3.3in]{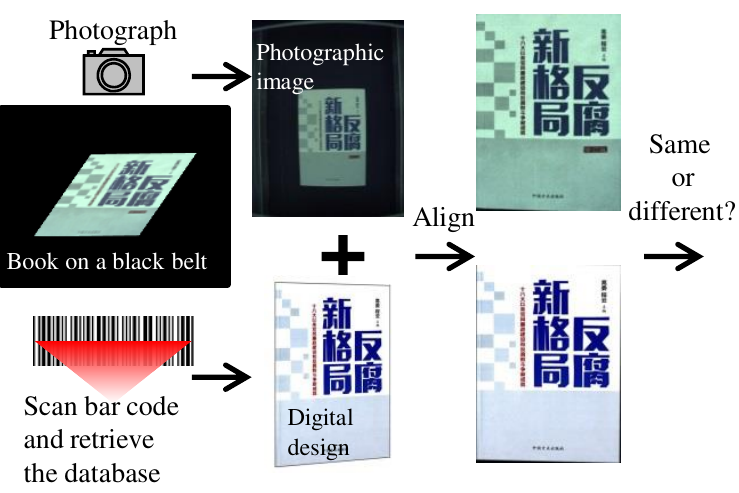}
	\caption{The pipeline of the application: The book is put on a black conveyor belt, and a camera on the top take picture of book cover. At the same time, the bar code at the back of the book is scanned, and the digital design of the book cover is retrieved form the database. After alignment, the algorithm judges if the photographic image and the digital design are same or different. }\label{fig:pipeline}
\end{figure}

Although changed book covers are rarely seen, it is necessary for a book seller to check if any real printed book cover is exactly the same as the digital design, which is very labor intensive and error-prone if done by human. So we design a device and propose an computer vision algorithm to save labor costs and improve accuracy. The structure of the device and the pipeline of the application is shown in Figure ~\ref{fig:pipeline}. The book is put on a black conveyor belt, and a camera is put on
the top to take picture of the book cover. At the same time, the bar code on the back of the book is scanned, and the digital design of the book cover is retrieved form the database. After alignment, our algorithm judges if the photographic image and the digital design are same or different.

Different from general verification task, which compares key features, this task values local features, as small changes in character or patterning mean the two images are different. For example, in face verification ~\cite{sun2015deeply}, faces with 10\% or more areas occluded are considered the same, but a book cover with a black area occlusion is considered different from its digital design. 
So the conventional verification methods are not suitable for this task. 
 
The task is similar to the game ``Spot the Difference'' \footnote{See wiki: https://en.wikipedia.org/wiki/Spot\_the\_difference}. However,  it is more complicated than the game, because the photographic image is interfered with illumination, color, and other noises. In other words, the same parts between two images are not exactly looked the same, and the criteria of ``difference'' need to be represented robustly. Subtraction of 
pixels dose not work well.

\begin{figure}
	\includegraphics[width=3.3in]{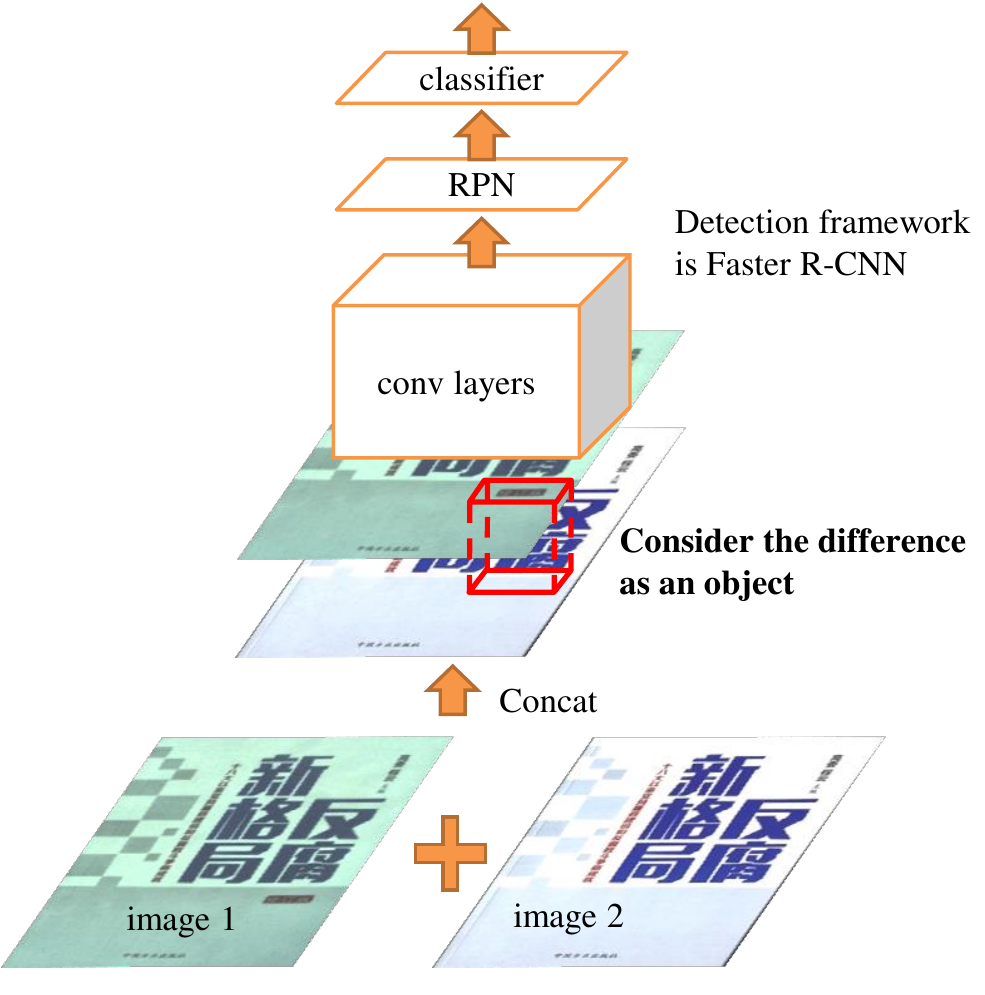}
	\caption{Spot the difference by object detection: two input images are concatenated along the channel dimension and the difference is considered as an object to detect. Faster R-CNN is adopted for object detection.}\label{fig:frame}
\end{figure}

To solve the spotting the difference problem, we propose a new method by CNN (Convolutional Neural Networks) based object detection. We stack the two RGB images as a 6-channel image, and consider the differences as objects to detect, then follow the pipeline of state-of-the-art object detection framework. In particular, the object detection framework is Faster R-CNN~\cite{ren2015faster}, as shown in Figure ~\ref{fig:frame}. We also propose an effective and efficient backbone network architecture
based on object detection network by stacking two branches of input data after several convolutional layers, which we call two-branch architecture.

To avoid costly data annotation, our training images are synthesized from weakly labeled image pairs without bounding boxes. The results show detection model trained on synthetic data performs well on both synthetic data and realistic data. 


In the experiments, we show that the object detection based method achieves much better performance than the general verification methods.  We also compress the network and observe that even with smaller and faster networks, out method can still achives good results. 


The rest of the paper is organized as follows. Section ~\ref{sec:related} briefly reviews the related work. Then algorithm details are introduced in section ~\ref{sec:spot}.
In section ~\ref{sec:experiments}, we conduct various experiments. Section ~\ref{sec:conclusion} concludes the paper.

\begin{figure*}
	\includegraphics[width=6.8in]{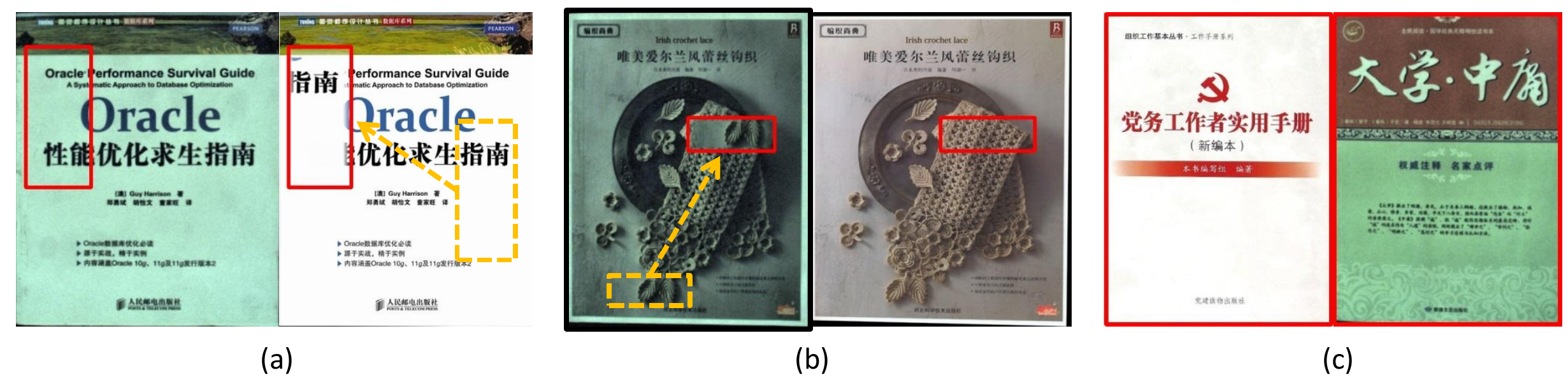}
	\caption{Three types of synthetic training data. For type (a), we sample a rectangle patch(in orange) from the digital design (in the right) and copy it to somewhere else(in red) as a difference from the photographic image (in the left) .  For type (b), we sample a rectangle patch from the photographic image and copy it to somewhere else as a difference from the digital design. For type (c), we sample miss-matched images as a training pair, of which the difference is the whole image.}\label{fig:traindata}
\end{figure*}

\section{Related Work}
\label{sec:related}

Many researches have been conducted in fields of image verification and change detection. However, most of these methods are not practical in this task. In this paper, we propose a new solution to this spotting the differences task, which  is related to object detection. 
In the following, we introduce similar task solutions and explain why they are not practical in this task, including verification and change detection. We also introduce methods we used -- object detection.

\subsection{Verification}
\label{sec:veri}
Verification problem has been developed relatively well in recent decades, including face verification ~\cite{chopra2005learning,taigman2014deepface,sun2015deeply} and signature verification ~\cite{plamondon1989automatic}.
The common outline of face verification consists of four stage: detect, align, represent, classify ~\cite{taigman2014deepface}.  In represent stage, feature extractor is designed to extract key features like structure of eyes and nose, but ignore other element like glasses, scars and wrinkle. In this spotting the differences task, there is no such key features, for every character and patterning matters. For example, a book cover image with a version number or without should be different. So the flow of verification can not solve it.

\subsection{Change Detection}

Change Detection problem contains lots of applications like video surveillance, remote sensing, civil infrastructure, etc ~\cite{radke2005image}.  However, different applications and image sources have different processing steps. For change detection in video surveillance, image sequence is calculated and the relationship of images in time series is important ~\cite{bianco2015far,wang2014cdnet}. In remote sensing application ~\cite{song2001classification, walter2004object, hussain2013change}, land-cover and land-use change information is considered, a core technique of state-of-the-art solution is doing segmentation and classification to images first, and then some comparison and analysis ~\cite{hussain2013change}. A solution for detecting changes of the three-dimensional structure of an outdoor scene is to estimate depth first and then compare the depth ~\cite{sakurada2013detecting}. One of the document image change detection methods uses OCR method to recognize characters in images first ~\cite{jain2013visualdiff}. Generally speaking, changes in different applications vary a lot and the solutions are not very practical in our task.

\subsection{Object Detection}

Numerous works have been proposed for object detection, and CNN based methods show leading results on  detection benchmarks. From R-CNN ~\cite{girshick2014rich}, many CNN based methods are proposed to improve both accuracy and speed, including Fast-R-CNN ~\cite{girshick2015fast}, Faster-R-CNN ~\cite{ren2015faster}, SSD ~\cite{liu2016ssd}, YOLO ~\cite{redmon2016you,redmon2016yolo9000}. 
CNN-based detection works for specific object like face ~\cite{qin2016joint} and text ~\cite{zhu2016scene} also show great advantage.
Different from hand-crafted feature extractor like SIFT ~\cite{lowe2004distinctive} and HOG ~\cite{dalal2005histograms}, features of a kind of object is learned from a deep neural network in these CNN based detection methods. Since the RGB pixel input can lead to features that describe a face or a text, it is reasonable that a pair of RGB pixel input or 6 channel input leads to features that describe ``difference''.

\section{Algorithm}
\label{sec:spot}


\subsection{Image Alignment}

The photographic image of book cover is taken on a black conveyor belt and the book cover is not at a fixed position of the image, thus we take a preprocessing pipeline to get a regular rectangle book cover, and then match it to the digital design.
Following is the processing details. 

First, the photographic image minus the pure black conveyor belt image background which still contains some texture. 
Then, we extract contour of book cover in the photographic image. We use Canny algorithm to extract edges and binary the image. After abandoning tiny pattern, we extract contour, approximate edges to a polygon, and finally extract a rectangle contour. 
Finally, we match and align the two images. 
We extract SIFT features from two images and match them, then filter the matches based on RANSAC and apply affine transformation to the digital design. 

	

\subsection{Training Data Synthetic}
\label{sec:simu}  In the image collection process, it may cost a lot to collect enough image pairs that have local differences to train a detection network. In other words, realistic data doesn't have enough data that contain ``objects''. On the one hand, changed book cover image pairs are in the minority, for in most cases, the book cover is not changed when printed. On the other hand, in this spotting the difference task, manually annotation makes neglect easily, as human eyes get tired soon.

To solve the lack of training data, we synthetic different book pairs from same ones, and the result shows the model trained on synthetic data performs well on realistic test set data. Some train set data is shown in Figure ~\ref{fig:traindata}. Synthetic details are as follows.

\begin{figure}
	\includegraphics[width=3in]{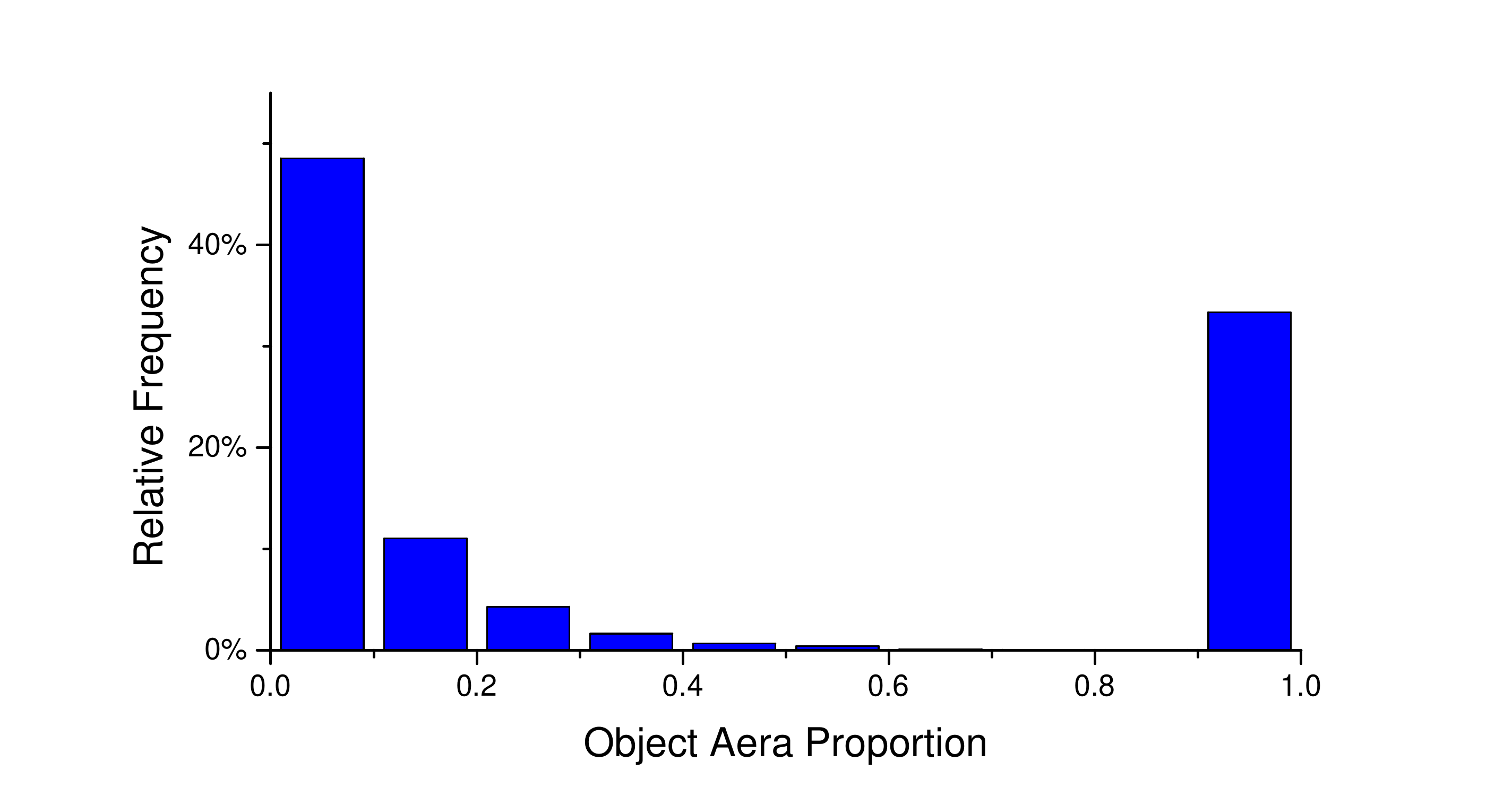}
	\caption{Object(difference) area distribution in synthetic training data. 
	Small differences which are less than 10\% of the image area occupies the most, and big differences as large as the whole image are the second.	}\label{fig:area}
\end{figure}

About 5k same image pairs are collected. We synthesize  different image pairs from these totally same pairs, of witch two-thirds are locally different and one third are globally different.

To synthesize locally different pairs, we choose a random image from a pair, either the photographic image or digital design. Then we sample a random rectangle patch in the selected image and copy it somewhere else in the same image. Some disturbances like margin blur and small-range scaling is implemented on the copied patch to make it seems nature. Thus, the new patch in the image is a difference or an object we need to detect.

We make sure the copied patch is different from the where it stuck to by simply compare the color histogram between the two patches. Otherwise, there would be many incorrectly labeled objects in the training data. After implementing the comparison of color histogram, there are still some wrongly labeled objects which are not easily discovered by the view of eyes, but the amount is significant reduced. In this way, tiny amount of noise data has little effect on the training.

For globally different image pairs, we choose a photographic image, then random choose miss-matched digital design, and resize it the same as the photographic image. These two images form a pair, and the object to detect is labeled as large as the whole image.

In this way, We synthesize only one object in every image pair, so we have a lot of objects or foregrounds to train, but backgrounds are not augmented more than the original 5k same image pairs. In order to make the model learn about backgrounds better, we still need to collect more data in the future.

We count the object size distribution as shown in figure ~\ref{fig:area}. 
Except for globally different pairs, we don't control the distribution of object area on purpose. We can see small objects which are less than 10\% of the whole image occupy the majority.
The detection model trained only on synthetic data generalizes well not only in synthetic validation data, but also on realistic data with various sizes of objects or multiple objects, as shown in Figure ~\ref{fig:goodresults}. It is noted that the synthetic validation data is not participated in the training process.

\subsection{Early-merging Architecture}
\begin{figure}
\includegraphics[width=3.2in]{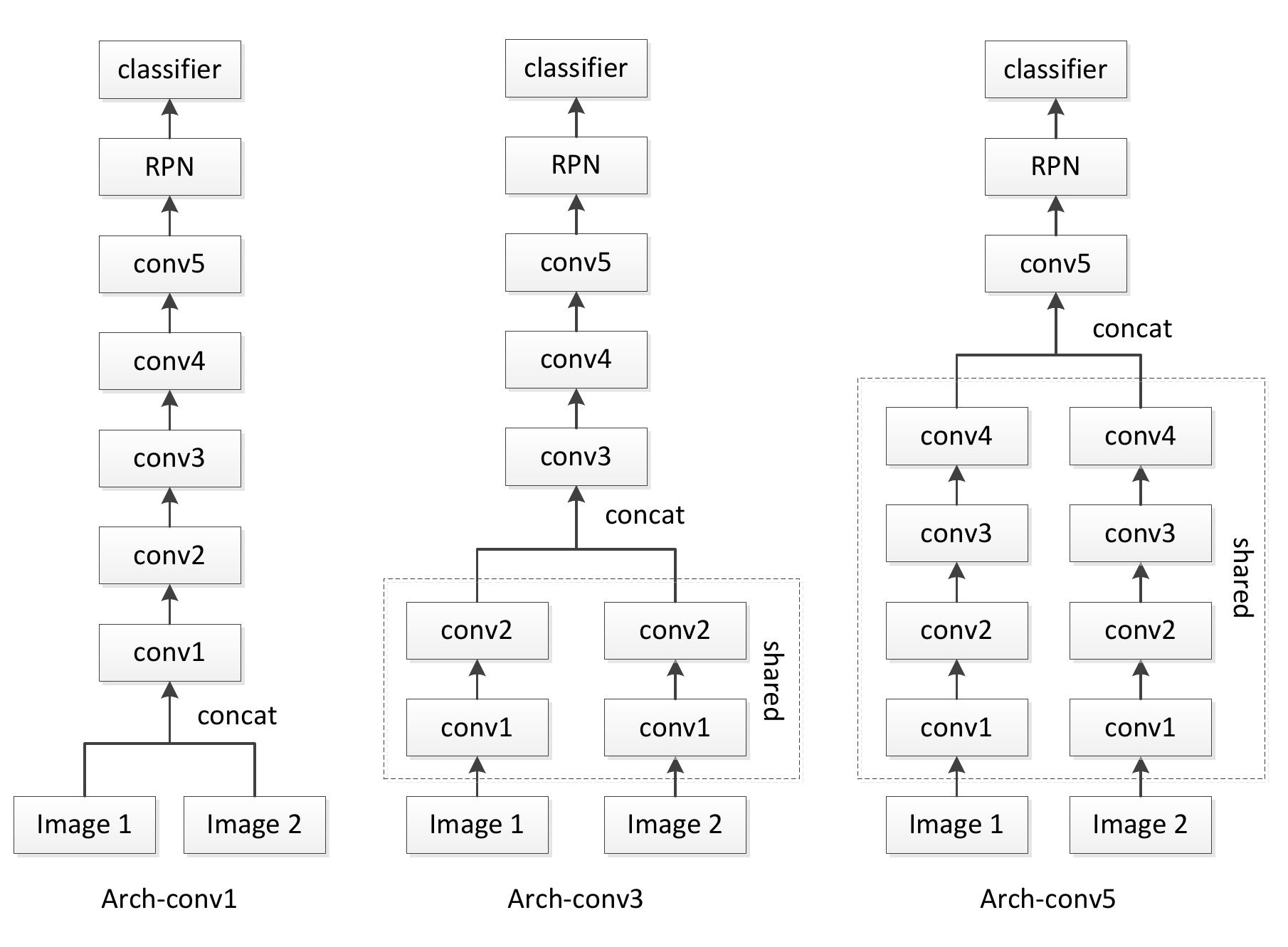}
\caption{Three architectures of our method. Conv1 to conv5 represent layer blocks in ZF net. ReLU, LRN and Pooling layers are omitted here. In the three architectures, two network branches are concatenated at different layers. Left: the Early-merging architecture. Middle: two branches are concatenated before conv3. Right: two branches are concatenated before conv5. }\label{fig:arch}
\end{figure}

The main idea of spotting the differences by object detection is to consider the two RGB 3-channel input images as one 6-channel image, and consider the differences as objects to detect. We use Faster R-CNN ~\cite{ren2015faster} as the detection pipeline, i.e., ``CNN feature extraction + region proposal + classification''. Input layer is modified to accept two images and two branches of data are combined by a Cancat layer. The idea is shown in Figure ~\ref{fig:frame}, and the
Early-merging network is shown in Figure ~\ref{fig:arch}. The CNN feature extraction network is ZF net~\cite{zeiler2014visualizing} which has five convolutional layers, and two image data layers are fed into a Concat layer immediately.

\subsection{Two-branch Architecture}
Considering the physical significance of two images, we design two-branches architecture. Two input images are fed into several Conv layers respectively, then are concatenated after several layers. Layers before the Concat layer share the same parameters. As shown in Figure ~\ref{fig:arch} middle and right, conv1 to conv5 represent layer blocks in ZF net, as other layers like ReLU, LRN and Pooling is omitted. Explain in detail, in arch-conv3, channels concatenate after pool2 and before conv3 layer; in arch-conv5, channels concatenate after relu5 and before conv5.

\section{Experiments}
\label{sec:experiments}

\subsection{Dataset and evaluation protocol}\label{sec:eval}
A sample in the dataset consists of two images -- a real book cover photographic image and it's digital design, and is labeled simple ``same'' or ``different'', without coordinate of bounding box. Realistic book cover image is photographed and then aligned with the digital design. Test set contains 1483 pairs of images, of which 982 pairs are same and 501 pairs different.  In the different set, 332 pairs are globally different and 169 pairs are locally different.

For evaluation, we consider same image pairs as ``positive'' and different pairs as ``negative''. The evaluation measure is \textit{Receiving Operating Characteristic Curve} (ROC Curve). As in practical applications , different pairs are more sensitive, which means if you mistakenly put the different pair as the same, it may cause big trouble like customer complaints;  on the contrary, if you mistakenly put the same pair as the different, it may only increase some labor cost. So in our evaluation, not only the \textit{area under ROC Curve} (AUC) is considered, but also the slope of the curve. 

\begin{figure}
	\centering
	\includegraphics[width=3.3in]{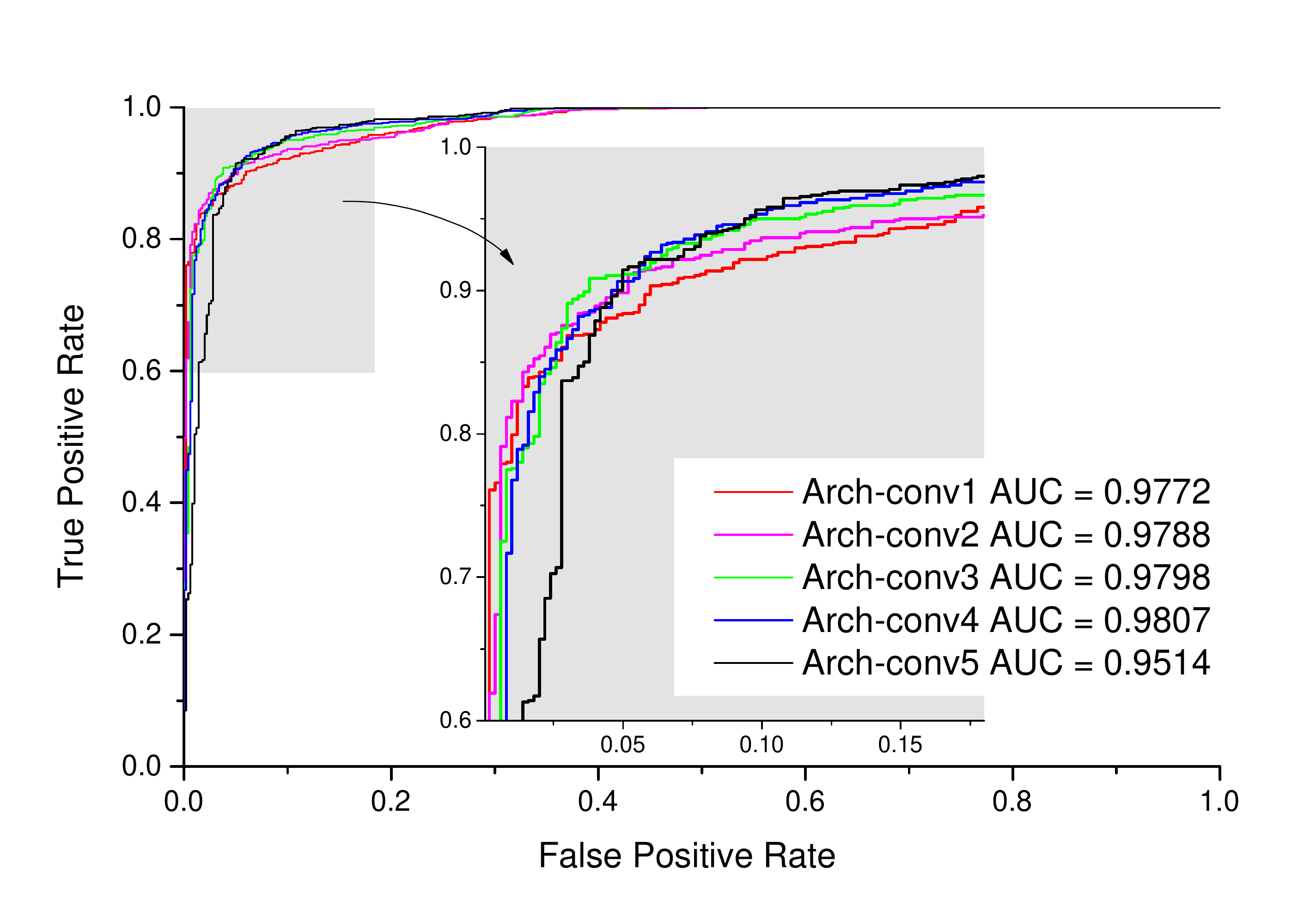}
	\caption{ROC Curves of five architectures. The label ``Arch-conv1'' represents two images are concatenated before conv1; ``Arch-conv2'' represents two branches are concatenated before conv2, and so on. }
	\label{fig:concatroc}
\end{figure}

\begin{figure*}
	\centering
	\includegraphics{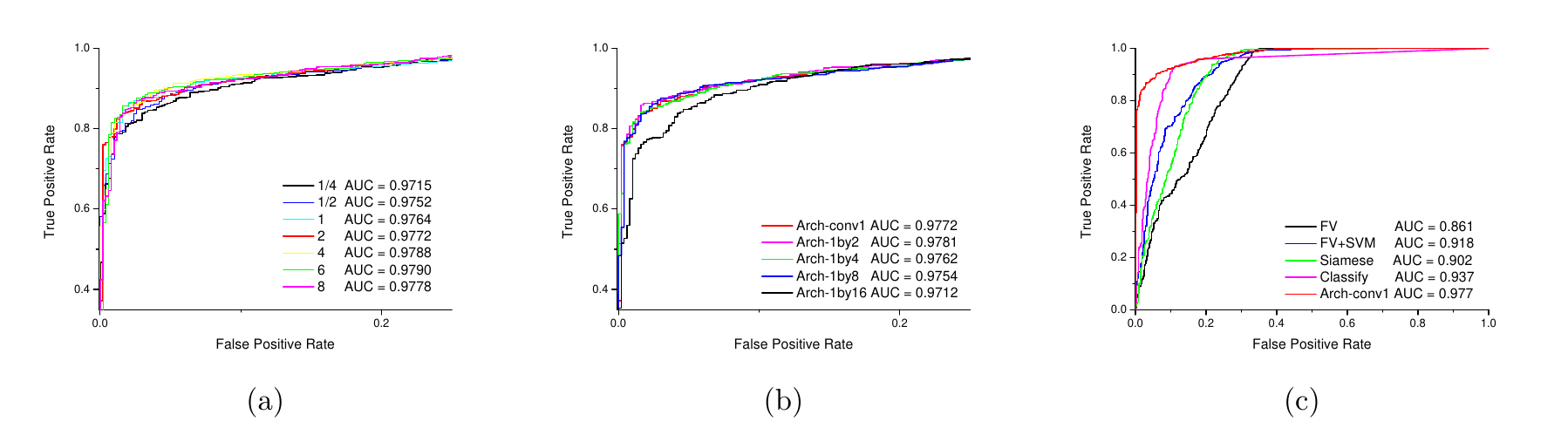}
	\caption{ROC Curves of three experiments.  (a)  Amount of synthetic training data.   (b) Simplified networks.  (c) Object detection method (Arch-conv1) compared with other methods. }
	\label{fig:combineroc}
\end{figure*}

\begin{figure}
	\includegraphics[width=3.3in]{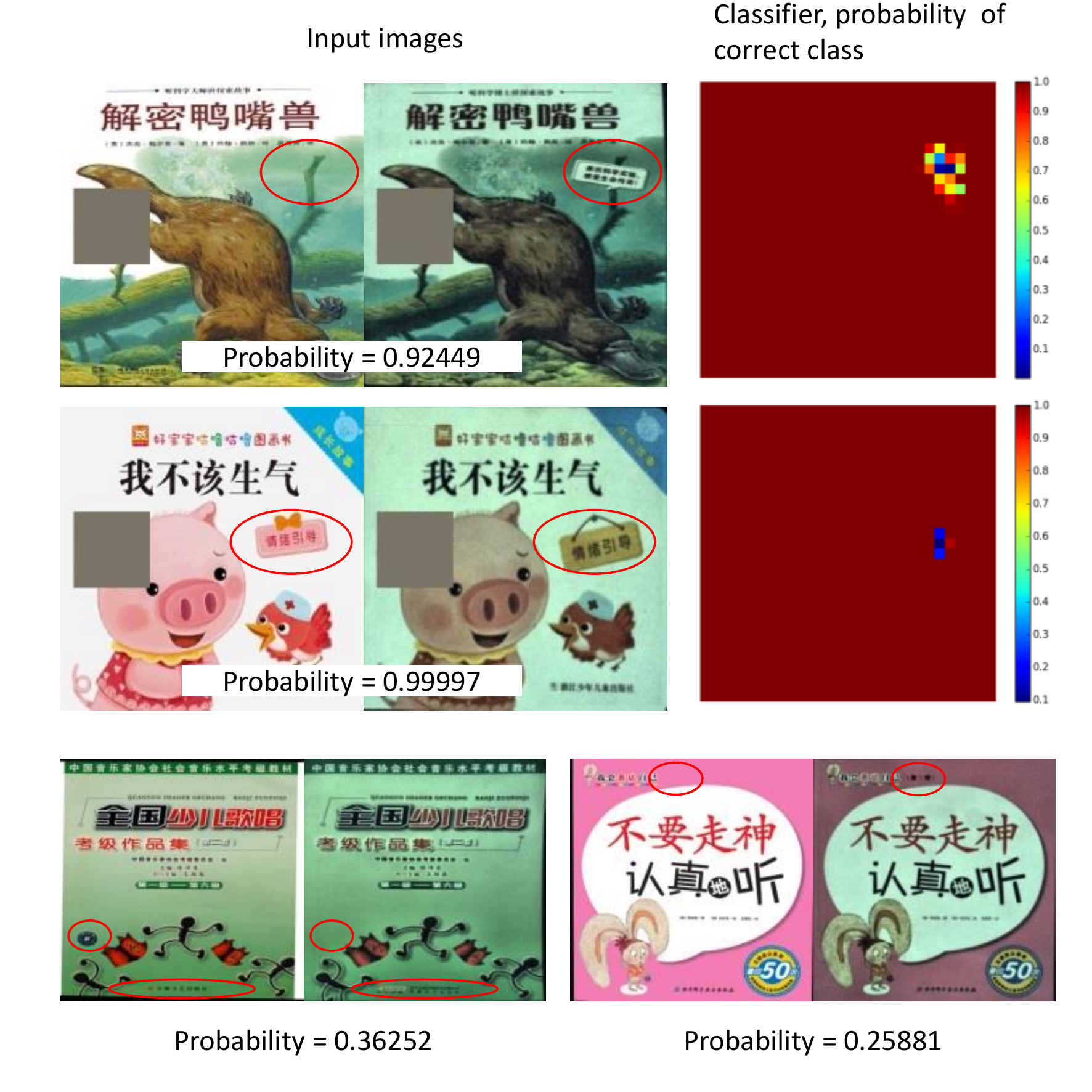}
	\caption{6-Channel classification network shows attention to the feature of ``differnce'' and predicts relative large differences. The differences are marked with red ellipses. The visualization is a map of probability as a function of the position of the gray square ~\cite{zeiler2014visualizing}. However, small differences are still ignored by 6-channel network as shown in the bottom. }\label{fig:6channel}
\end{figure}

\subsection{Object Detection Method}

\subsubsection{Training details} We adopt Faster R-CNN~\cite{ren2015faster} as the object detection framework, with ZF net ~\cite{zeiler2014visualizing} as CNN feature extraction layers. Two images are inputted, and  then fed into a Concat layer immediately or after several layers (Figure ~\ref{fig:arch}). We initialize Conv layers before Concat layer by pretrained model from ImageNet, and Conv layers after the Concat layer with the ``xavier'' method ~\cite{glorot2010understanding}. It is noted the experiment results show it doesn't help to initialize layers after Concat layer by pretrained model from ImageNet. 
In both training and testing, we use single-scale method, where the image's shorter side is \begin{math} s=600\end{math} pixels. The horizontally flipping is implemented as training data augmentation. 

The network is trained end-to-end as describe in ~\cite{ren2015faster} as approximate joint training. The hyper-parameters for training Faster R-CNN are almost the same in \cite{ren2015faster}.  We set global learning rate as 0.001 at the beginning. After 10 epochs, we lower the learning rate to 0.0001 to train for 4 epochs.

In evaluation, as the testing set images are labeled binary ``same'' or ``different'', we consider the highest object score as the distance between the two images. In other words, the larger object score means higher possibility the two images are different, lower score means the two images are same. Then we consider same image pairs as ``positive'' and draw ROC Curve to analysis as described in section ~\ref{sec:eval}. 

\subsubsection{Amount of synthetic training data}
There are about 5k same image pairs in total. We synthesize training data as described in ~\ref{sec:simu}.
We experiment on different amount of synthetic training data to explore if more synthetic training data is better.
The result is shown in Figure ~\ref{fig:combineroc} (a). The tag means synthetic magnification(TODO:check) to the original 5k image pairs, as ``1/4'' means the amount of synthetic training data is one fourth of the 5k image pairs, ``2'' means twice as the 5k image pairs, and so on. 

The ROC Curve shows that there is a tendency that more synthetic training  data leads to better test performance, but when the magnification larger than 1, the increase becomes slow, even become worse when magnification larger than 6. The result shows it helps a little to synthetic training data more than the original 5k same image pairs. The reason is that 
we can synthesize infinite objects if we want, but the amount of background can not be expanded beyond these 5k pairs. In order to make the model learn about backgrounds better, we still need to collect more data in the future.

\subsubsection{Two-branch Architecture}\label{sec:twobranch}  
We experiment different architectures that two branches are concatenated after several layers. More than shown in Figure ~\ref{fig:arch}, we experiment five architectures, trying all the Concat layer locations before five Convolution Layers. The amount of synthetic training data is twice the original 5k. 

The performance of the five architectures is in Figure ~\ref{fig:concatroc}. The AUCs of five architectures are fairly close, among which Arch-conv4 has a small advantage. But if we look into the slope of the curve, we find the Arch-conv1(Early-mearing architecture) close to the \begin{math}y\end{math} axis most, which means in lower false positive rate, Arch-conv1 can reach higher true positive rate. In other words, if the  application prefers to rarely miss different image pair, Arch-conv1 may be the best one. 

Besides, Arch-conv1 has the least parameters and least computational cost, for layers with same Convolution Layer name in all the architectures have same channels, except for the Concat layer doubles the channel. 

Table ~\ref{tab:param} shows the parameters and theoretical computational cost, which is given as the number of adds and multiplications(MAC) of five architectures. As the architectures are exactly the same from conv5 to the end, we only count the number of parameters and MAC from beginning to conv5. The size of input image is assumed as 1000*600.

\begin{table}
    \small
    \center
	\caption{Params and MAC of architectures}
	\label{tab:param}
	\begin{tabular}{ccl}
		\toprule
		Architecture&Params&MAC\\
		\midrule
		Arch-conv1 & 3.74M & 17.9G \\
		Arch-conv2 & 4.35M & 23.82G \\
		Arch-conv3 & 5.24M & 26.04G \\
		Arch-conv4 & 6.57M & 29.35G \\
		Arch-conv5 & 7.45M & 31.56G \\
		Arch-1by2 & 941.86k & 5.55G \\
		Arch-1by4 & 239k & 1.92G \\
		Arch-1by8 & 61.52k & 748.22M \\
		Arch-1by16 & 16.16k & 320.93M \\
		\bottomrule
	\end{tabular}
\end{table}

This result is consistent with ~\cite{zagoruyko2015learning}, of which experiments show good flexibility when relations between two images are learned from the beginning convolutional layer.

\subsubsection{Model Simplification} As we consider the difference between two images as an object, it is an object with much simpler features, compared to objects like human, cat, vehicle, etc. In this situation, it is possible that simpler network with fewer parameters still has a good performance. We tried some simplified networks. 

Based on Arch-conv1, we simply reduce channel of layers from beginning to conv5,  rpn\_conv, fc6 and fc7 to half (Arch-1by2) or quarter (Arch-1by4), and so on. Training details are the same as above.
The ROC Curves (Figure~\ref{fig:combineroc} (b)) show that these simplified networks perform quite close to the Arch-conv1, until we reduce the channel to one-sixteenth there appears a relatively large decline. 

The parameters and MAC of these simplified architectures are shown in Table ~\ref{tab:param}. It is noted that the layers after conv5 are not counted.  
Compared to Arch-conv1, the Arhc-1by8 has about 24 times fewer MAC, but still achieves close performance on test set data. 

\begin{figure*}
	\includegraphics{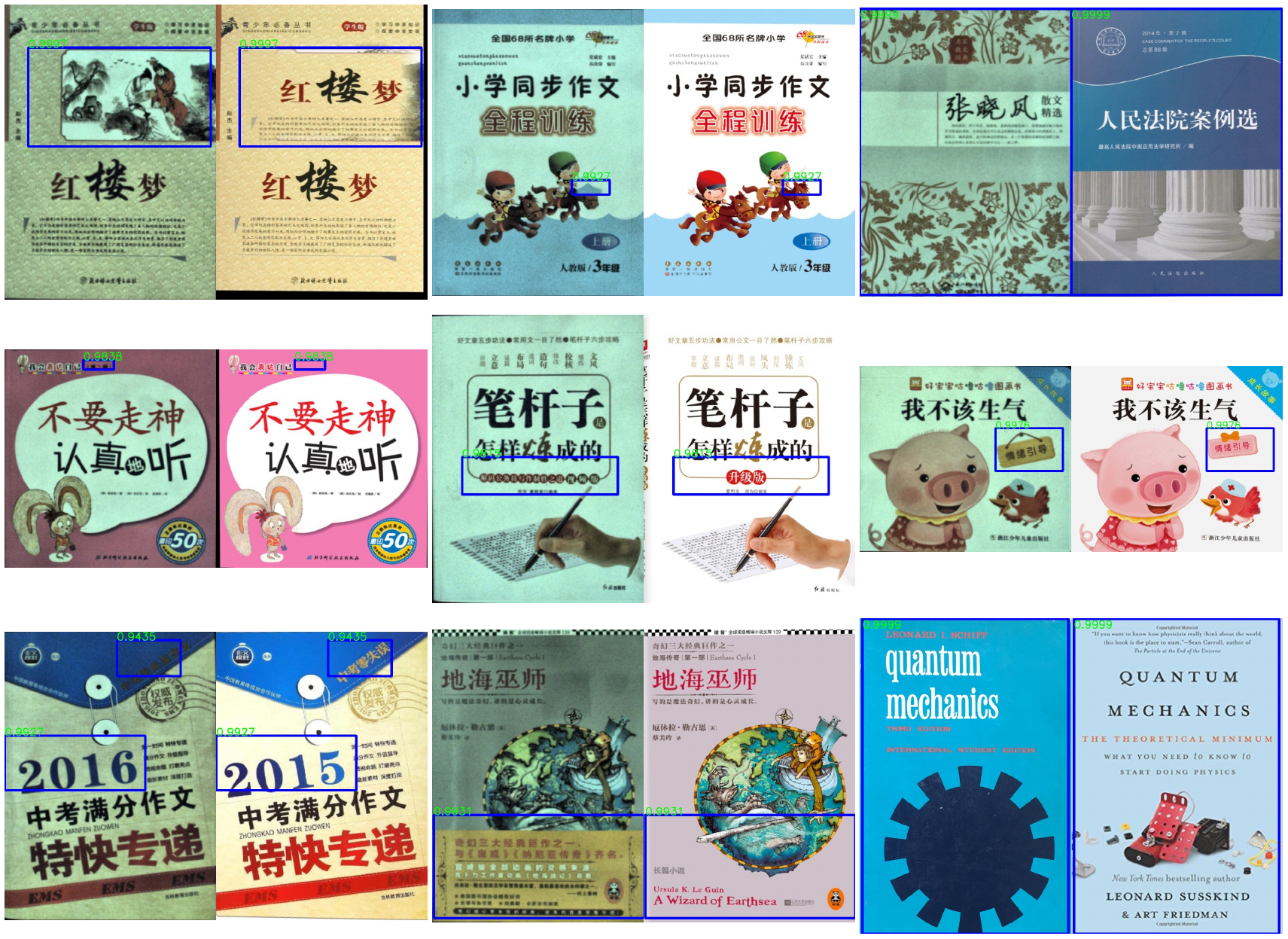}
	\caption{Some qualitative results. We can see the detection model trained on synthetic data performs well on differences of character, patterning, book girdle, and global design, no matter single or multiple objects in a pair. The samples on first row are synthetic validation data and others are realistic data. }\label{fig:goodresults}
\end{figure*}

\subsection{Comparison with other methods}
The evaluation compares the object detection method with four other baselines. The methods are described in details below and we analysis at the end.


\subsubsection{Fisher Vector} Before deep learning is widely used in image recognition, one of  common methods is to encode dense features of an image to a single feature vector, and the vector represents the image. Fisher Vector(FV) encoding has been used in image recognition tasks and had a good performance ~\cite{sanchez2013image,simonyan2013fisher}. Following a classic recognition example in ~\cite{vedaldi2012vlfeat}, we extract dense SIFT feature and encode it to a feature vector by FV, then compare Euclidean distance between two feature vectors of two images. 

\subsubsection{FV + SVM} After extract feature vectors of the two images by FV encoding, which is the same as the above, we carry out a SVM as diagonal metric learning method. We use a conventional linear SVM, and the feature fed in SVM is the vector of squared difference between the two compared FVs.  As for train set data, we use all the 5k same image pairs we have collected and the same amount of different image pairs. Different images pairs includes about 1k realistic different image pairs, and the rest is random selected in all the synthetic training data described in ~\ref{sec:simu}.

\subsubsection{Siamese Architecture } The Siamese architecture is widely used in object verification tasks, such as face verification ~\cite{chopra2005learning}, signature verification ~\cite{bromley1993signature}. We use ZF net as parameter-shared convolutional network. 4096-dimension vector is extracted and fed into the loss function of \textit{ContrastiveLoss}\footnote{http://caffe.berkeleyvision.org/tutorial/layers/contrastiveloss.html}. The train set data is the same as the FV + SVM described above.

\subsubsection{6-Channel Classification} In ~\cite{zagoruyko2015learning}, to compare similarity of a pair of images, the input layer is a 2-channel image which is stacked by two gray-scale images, then followed a series of convolutional layers, and finally a fully connected linear decision layer. Similar to ~\cite{zagoruyko2015learning}, we concatenate two 3-channel images as a 6-channel image and feed it to a ZF classification network.

\subsubsection{Analysis}Figure ~\ref{fig:combineroc} (c) shows the ROC Curves of different methods. The curve of 6-channel classification method is tagged as ``Classify''.  As we can see, our method has a comprehensive advantage. 

Verification methods including FV, FV + SVM and Siamese can find out globally different image pairs precisely, but can not distinguish locally different image pairs. 
As we describe in Section ~\ref{sec:veri}, in verification solution outline, features are extracted to represent the image and then fed into a classifier.  
In Fisher Vector method, hand-crafted features are extracted and encoded, and Euclidean distances between two feature vectors represents the non-similarity between two images. Distance representation is learned in FV + SVM. In Siamese method, not only distance representation but also image feature is learned by training. 
These methods consider features among an image globally, and a small locally pattern's feature is extracted indifferently among the whole image. So the comparison of a global feature vector can not distinguish image pairs that have slightly local differences.


6-Channel classification shows relatively large advantage. The relation between two images are leaned from the beginning, and we can see the classification network shows attention to the feature ``difference'' as shown in Figure  ~\ref{fig:6channel}. However, small but important differences are still ignored by the 6-channel network.

On the contrary, our method using object detection method to consider local information. The feature of ``difference'' is learned via CNN, and regions of possibly local differences are proposed by Region Proposal Network in Faster R-CNN. So in our object detection method,  features of locally differences will not fade away in a whole image.




\begin{figure*}
	\includegraphics{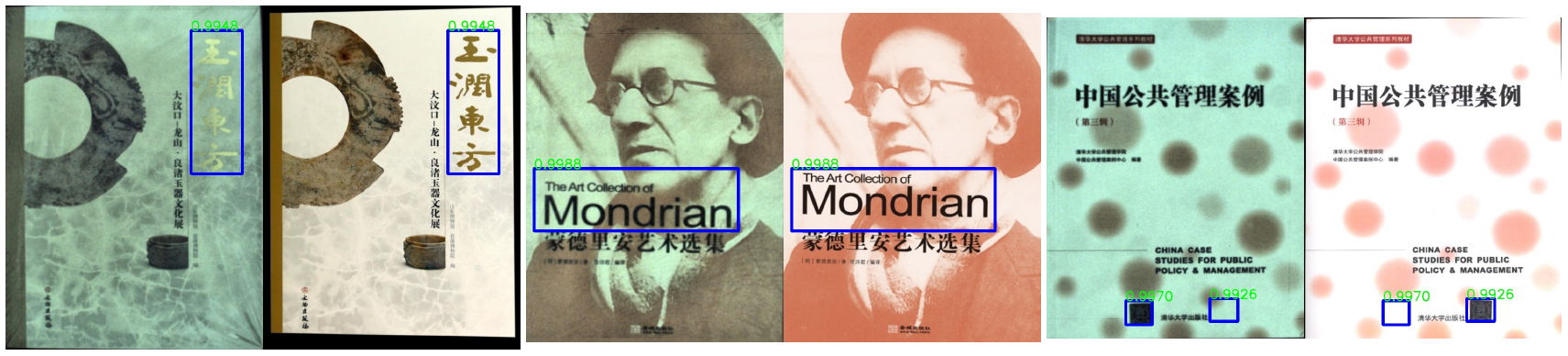}
	\caption{Some false detection examples. In the left pair, some characters are printed in reflecting material; In the middle, a piece of characters moves a small distance; In the right, an icon is not printed but pasted on the book, which has not exactly the same pose or location as in the digital design. Technically, there are some differences in visual view, but in business application these differences are not important. }\label{fig:badresults}
\end{figure*}

\begin{figure*}
	\includegraphics{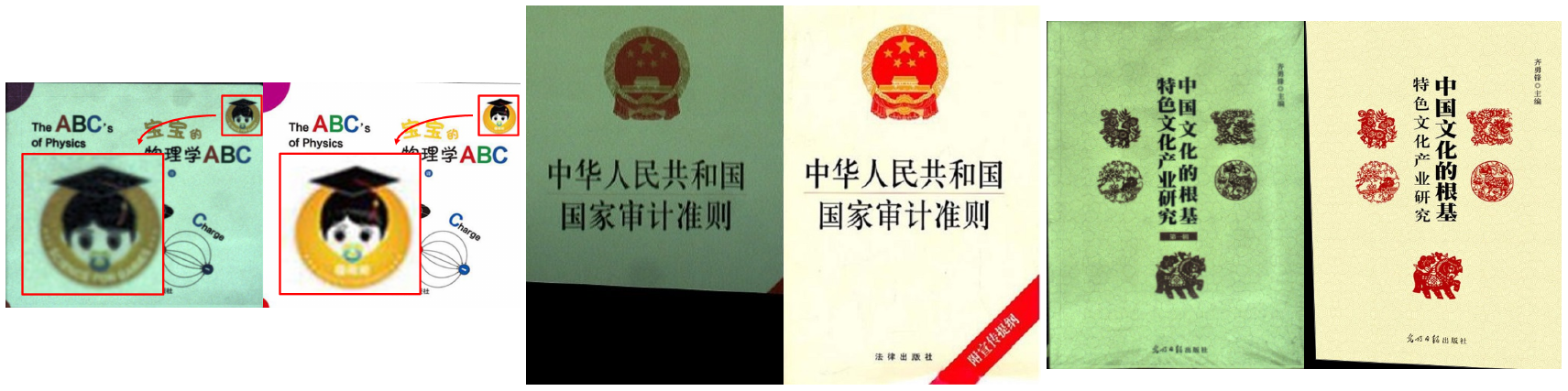}
	\caption{Some missed detection examples. In the left pair, the difference is too unapparent. In the middle, the reason is that the black margin is trained as background. In the right, two images are not aligned well.  }\label{fig:leaveresults}
\end{figure*}

\section{Discussion}

Figure~\ref{fig:goodresults} shows some examples of qualitative results from Arch-conv1. We can see, the differences including characters and patterings can be detected  precisely. Besides, the noise is ignored , including illumination, color, noisy points and slight stains. This is because these kinds of noises are treated as background in the training data. Correspondingly, the real differences, or objects, are treated as foreground. In the training processing, features of foreground are learned, as well as background.

However, there are also some false or missed detection. Figure ~\ref{fig:badresults} shows some false detection examples. Actually, in visual point of view, there are some differences between two images, but in application they should be divided into the noise category. In the left, some characters are printed in reflecting material, so the photographic image looked quite different form the digital design. In the middle, a piece of patterning moves a small distance. Technically, this is a kind of difference, but is not very important in selling books. Similarly, in the right, an icon is not printed but pasted on the book, which has not exactly the same pose or location as in the digital design.

In these false detection cases,  reflecting printing, small movements and stuck icons should be some kind of background, similar to illumination or color noise, but actually they are closer to differences to detect.  At present, these cases account for a tiny proportion in the whole dataset. If they cause big problem in the future, which means there would be enough data of these cases, we have an improvement idea, that is, to treat the reflecting printing, small movement and stuck icon as other kinds of objects. In this way, there would be four classes of objects to detect, including real difference.

Some image pairs do have differences, but are not detected, as shown in Figure~\ref{fig:leaveresults}. In the left, the difference is too unapparent that even human eye can't 
find it easily. In the middle, the photographic image is somehow not completely and missed part is filled with black pixel. This case is not detected, because a lot image have black margins and the black margin is trained as  background. In the right, the difference between two images is quite obvious but has a low detection score. The reason is that the two images are not aligned well, and this kind of data is fairly few in training set. 

Besides verifying and detecting the changes between book cover digital design and its photographic image, our proposed method can generalize to similar applications, such as document change, places change, object change and so on. 

\section{Conclusion}
\label{sec:conclusion}
We have presented an object detection method for the task of detecting changes between two images.  
Our method is fast, accurate and robust while using very cheap annotation.
The method stacks two 3-channel images as one 6-channel image and treats the differences as objects. Faster R-CNN is used to detect the differences.
We design the Early-merging architecture that merges two branches of image data at the beginning of the network. Compared with other architectures, Early-merging performs better and faster by sharing convolution computation between the two branches.
Besides, our method is robust. After model compression, our method can run 24 times faster while keeping the accuracy. 
To avoid expensive annotation, all the detection training data are synthesized from the weakly labeled image pairs without bounding boxes. Even so, the trained model can spot the difference effectively on both synthetic data and realistic data. 
We verify the proposed method on a real task that detects the difference between the digital design and photographic image of a book. 
Compared with the verification based methods, including Fisher Vector coding, Siamese, and 6-channel classification, our detection based method achieves the best performance.

{\small
\bibliographystyle{ieee}
\bibliography{sigproc} 
}

\end{document}